\documentclass[]{spie}  %>>> use for US letter paper
\usepackage{amsmath,amsfonts,amssymb}
\usepackage{graphicx}
\usepackage[colorlinks=true, allcolors=blue]{hyperref}
\usepackage{printlen}
\usepackage{multirow}
\usepackage{placeins}

\title{Fast emulation of density functional theory simulations using approximate Gaussian processes}

\author[a]{Steven Stetzler}
\author[b]{Michael Grosskopf}
\author[b]{Earl Lawrence}
\affil[a]{University of Washington Department of Astronomy, 3910 15th Ave NE, Seattle, WA, USA}
\affil[b]{Los Alamos National Laboratory, PO Box 1663, Los Alamos, NM, USA}

\authorinfo{Further author information: (Send correspondence to S.S.)\\S.S.: E-mail: stevengs@uw.edu}

\begin{document} 
\maketitle

\begin{abstract}
Fitting a theoretical model to experimental data in a Bayesian manner using Markov chain Monte Carlo typically requires one to evaluate the model thousands (or millions) of times. When the model is a slow-to-compute physics simulation, Bayesian model fitting becomes infeasible. To remedy this, a second statistical model that predicts the simulation output -- an ``emulator'' -- can be used in lieu of the full simulation during model fitting. A typical emulator of choice is the Gaussian process (GP), a flexible, non-linear model that provides both a predictive mean and variance at each input point. Gaussian process regression works well for small amounts of training data ($n < 10^3$), but becomes slow to train and use for prediction when the data set size becomes large. Various methods can be used to speed up the Gaussian process in the medium-to-large data set regime ($n > 10^5$), trading away predictive accuracy for drastically reduced runtime. This work examines the accuracy-runtime trade-off of several approximate Gaussian process models -- the sparse variational GP, stochastic variational GP, and deep kernel learned GP -- when emulating the predictions of density functional theory (DFT) models. Additionally, we use the emulators to calibrate, in a Bayesian manner, the DFT model parameters using observed data, resolving the computational barrier imposed by the data set size, and compare calibration results to previous work. The utility of these calibrated DFT models is to make predictions, based on observed data, about the properties of experimentally unobserved nuclides of interest e.g. super-heavy nuclei.
\end{abstract}

% Include a list of keywords after the abstract 

\section{INTRODUCTION}
\label{sec:intro}  % \label{} allows reference to this section

Computer simulation of complex phenomena plays a key role in modern science and engineering. Increased computational power has led to improved simulation fidelity, however the time-cost of the best models still limits their use. For instance, to do optimization or uncertainty quantification with a simulation, hundreds or even many thousand evaluations may be required\cite{kennedy2001bayesian,higdon2004combining, emulation_higdon_2008}. Emulators, also commonly referred to as surrogate models, are used as fast approximations to the expensive simulation in order to facilitate their use in contexts where many evaluations are required. 

Data-driven emulators use a limited number of evaluations of the simulator and learn to predict new outputs at unseen input locations. For applications like uncertainty quantification, it is often critical that the emulator gives a measure of uncertainty with the prediction, reflecting the credible values of the simulator at the unseen location. Gaussian process (GP) models are a common approach used for emulation\cite{kennedy2001bayesian,higdon2004combining}. GPs are a form of Bayesian regression, representing a probability distribution on unknown functions and using Bayesian updating to refine this distribution with observations from the simulator\cite{rasmussen2003gaussian}. They provide a flexible, non-linear predictor with uncertainty that is well-suited for emulation problems. 

Gaussian processes main drawback is poor scaling with increasing number of observations, $n$. The computational cost scales as $n^3$ and memory use scales as $n^2$ due to the need to build and invert a covariance matrix between all observations. This makes the use of GPs infeasible for more than a few tens of thousands of observations. Several approximate Gaussian process methods have been proposed to reduce this cost. In this work, we focus on one class of approximate Gaussian process – sparse, variational, inducing-point based methods -- the sparse Gaussian process (SGP)\cite{sparse_gp_titsias} and the stochastic, variational Gaussian process (SVGP)\cite{hensman2013gaussian,svgp_hensman}. Because inducing-point methods reduce the expressive capability of the Gaussian process, we also include deep kernel learning (DKL) Gaussian processes\cite{deep_kernel_wilson}. DKL uses a neural network to learn a non-linear transformation of the inputs that then feeds into the Gaussian process covariance function. This transformation allows the inducing-point GP to find a latent representation of the inputs that improves its predictive capability. We pair DKL with the SGP to create the deep kernel sparse Gaussian process (DKSGP) and with the SVGP to create the deep kernel stochastic variational Gaussian process (DKSVGP).

We explore the performance of these approaches to emulation of a nuclear density functional theory (DFT) model, UNEDF1. Nuclear DFT allows for simulation of heavy nuclei that provides important insights for other sciences, including r-process nucleosynthesis in nuclear astrophysics \cite{mumpower2016impact}. The functionals include parameters that must be compared to experimental observations in order to provide a calibrated model for prediction and future application. Being able to quickly approximate the output of the UNEDF simulations can allow for parameter calibration, prediction with uncertainty, and comparison between parameterizations of the functionals. For this work, we will focus on one parameterization, UNEDF1\cite{PhysRevC.85.024304, schunck2020calibration}. Refs.~\citenum{PhysRevC.85.024304} and \citenum{schunck2020calibration} provide more details on the physics in the DFT simulation. In this work, we focus on emulating the binding energies predictions of the UNEDF1 model for 75 nuclides with 500 different parameter settings. Treating the binding energy prediction as a univariate output of the simulation with proton number, $Z$, and neutron number, $N$, as parameters generates a dataset size of $n = 37,500$. A GP emulator cannot be used straightforwardly with a data set of this size. Previous work has used a multivariate emulator approach, treating the UNEDF1 model output at each parameter setting as a $75$-dimensional vector quantity. This allows the dimensionality of the emulator input to reduce to $n = 500$, the number of different parameter settings \cite{mcdonnell2015uncertainty,higdon2015bayesian,schunck2020calibration}. Reducing the size of the input data for the Gaussian process restores computational feasibility but reduces the capability of the emulator. Most importantly, a multivariate emulator built for 75 nuclide binding energies is unable to predict the binding energy for any other nuclide. Rather, it would require building a new emulator after evaluating the simulator on the new nuclide for all parameter combinations. In this work, we treat Z and N as inputs to the approximate GP allowing for prediction of the binding energy for new nuclides for calibration, optimization, and other emulator applications. Treating Z and N as inputs returns the problem to $n = 37,500$, necessitating the use of an approximate-GP based emulator.

In the following, we will compare the performance of approximate GP emulators, along with the multivariate emulator, on the UNEDF1 simulation data. Section \ref{sec:background} provides background information about Gaussian processes and the approximate GPs in use. Section \ref{sec:experiment} discusses the UNEDF1 simulation data as well as the design of our experiments with the approximate GPs. Section \ref{sec:training} outlines our process for training the emulators i.e. selecting GP hyperparameters. Section \ref{sec:results} discusses the performance of the approximate GPs in prediction of simulation outputs (Sec.~\ref{sec:evaluation}) as well as when used for inference of the UNEDF1 simulation parameters that best predict the observed binding energies of 75 nuclides in a maximum likelihood sense (Sec.~\ref{sec:optimize}) and in a Bayesian inference sense (Sec.~\ref{sec:calibration}).

\section{BACKGROUND AND METHODS}
\label{sec:background}

We provide a brief introduction to Gaussian process regression in Sec.~\ref{sec:gp}, approximate versions of the Gaussian process in Sec.~\ref{sec:approx_gps}, and describe their implementation in Sec.~\ref{sec:gpytorch}.

\subsection{Gaussian Process Regression}
\label{sec:gp}

The Gaussian process is one approach to the modelling of observed targets $y = [y_1, ..., y_n]^T$ with $y_i \in \mathbb{R}$ as the output of an unknown function $f$ evaluated at inputs $X = [x_1, ..., x_n]^T$ with $x_i \in \mathbb{R}^d$:
\begin{align*}
    y = f(X).
\end{align*}
The Gaussian process places a prior on the distribution of functions $f$:
\begin{align*}
    f \sim \mathcal{N}(\mu(X), \Sigma(X, X; \theta)),
\end{align*}
a multivariate normal distribution where $\mu$ is the process mean and $\Sigma$ is the process covariance matrix. The covariance matrix is built using a ``kernel'' function $k(x_i, x_j; \theta)$ defined between two points $x_i, x_j \in \mathbb{R}^{d}$ which typically depends on one or more hyperparameters $\theta$. The covariance matrix between $n$ points in $X$ is evaluated as:
\begin{align}
    \Sigma(X, X; \theta) &= \left[ k(x_i, x_j; \theta) \right]_{1 \leq i,j \leq n}.
\end{align}
The kernel function satisfies the property of an inner product, namely that it is symmetric and produces values greater than $0$. These properties ensure the covariance matrix is positive definite, a necessary requirement for defining a multivariate normal. The log-likelihood of observations $y$ under the GP prior is:
\begin{align}
    \label{eqn:L_Exact}
    \begin{split}
    \mathcal{L}(y | X, \theta) &= \log \mathcal{N}\left(y | \mu(x), \Sigma(X, X; \theta)\right) \\
    &= -\frac{1}{2} \left((y - \mu(X))^T \Sigma(X, X; \theta)^{-1} (y - \mu(X)) + \log \left(2\pi^d \vert \Sigma(X, X; \theta) \vert\right)\right).
    \end{split}
\end{align}
Importantly, this log-likelihood depends on the choice of kernel hyperparameters $\theta$. Kernel hyperparameters are typically chosen by maximizing the log-likelihood through an optimization method. Prediction at new input points is performed by conditioning the GP prior on the training data, creating a predictive distribution that is still a multivariate normal with a new mean and covariance. At a new point $x^* \in \mathbb{R}^{d}$, the GP predicts a distribution over target values $y^*$:
\begin{align}
    \label{eqn:pred}
    \begin{split}
    y^{*} &\sim \mathcal{N}\left(\tilde{\mu}, \tilde{\Sigma}\right) \\
    \tilde{\mu} &= \mu(x^*) + \Sigma(x^*, X) \Sigma(X, X)^{-1} \left(y - \mu(X) \right) \\
    \tilde{\Sigma}(x^*, x^*) &= \Sigma(x^*, x^*) - \Sigma(x^*, X)\Sigma(X, X)^{-1}\Sigma(X, x^*)
    \end{split}
\end{align}
Evaluations of the log-likelihood and predictions at new input points require inverting the $n \times n$ covariance matrix $\Sigma(X, X)$, an operation that takes $\sim n^3$ operations. This is the primary restriction of the application of GP regression to large data sets with $n \gtrsim 10^4$.

\subsection{Approximate Gaussian Processes}
\label{sec:approx_gps}
 
One way to speed up Gaussian process regression is to use inducing point methods. Inducing point methods use $m$ pseudo-points in the input space to provide predictions that approximately match their exact counterpart. These points can be chosen from the input data or their locations learned by using an optimization method. In our notation, the inducing points are treated as a matrix $X_m = [x_1, ..., x_m]^T$ with $x_i \in \mathbb{R}^{d}$. Speedup is obtained with these methods when $m < n$ and typically $m$ is chosen so $m << n$. Ref.~\citenum{sparse_gp_titsias} introduces the sparse variational GP (SGP) and derives a variational lower bound of the log-likelihood (Eqn.~\ref{eqn:L_Exact}), the evidence lower bound (ELBO), when treating the inducing points as variational parameters:
\begin{align}
    \label{eqn:L_SGP}
    \begin{split}
    \mathcal{L}(y \vert X, \theta) &\geq \mathcal{L}_{SGP}(y \vert X, \theta) \\
    &= \log\left[\mathcal{N}(y \vert \mu, \Sigma(X, X_m) \Sigma(X_m, X_m)^{-1} \Sigma(X_m, X))\right] \\
    &+  \frac{1}{2\sigma^2} Tr\left(\Sigma(X, X) - \Sigma(X, X_m) \Sigma(X_m, X_m)^{-1} \Sigma(X_m, X)\right).
    \end{split}
\end{align}
The locations of the inducing points are treated as additional hyperparameters and are chosen to maximize Eq.~\ref{eqn:L_SGP}. This bound converges to the exact value (Eq.~\ref{eqn:L_Exact}) when the inducing points approach the input data in both number and location. Ref.~\citenum{svgp_hensman} introduces the stochastic variational GP (SVGP) by parameterizing the variational distribution over the inducing point values $u$:
\begin{align*}
    q(u) &= \mathcal{N}\left(u \vert \mu_{u}, \Sigma_{u} \right),
\end{align*}
parameterized in terms of a mean vector $m$ and covariance matrix $S$. This produces an updated ELBO given in Eq.~4 of Ref.~\citenum{hensman2013gaussian}. 

The inducing point mean $\mu_{u}$ and covariance $\Sigma_{u}$ are learned as additional hyperparameters, and the entries are chosen to maximize the ELBO. The ELBO calculation decomposes into a sum of contributions from individual training points, allowing gradients of the entries of $\mu_{u}$ and $\Sigma_{u}$ with respect to the training data to decompose as a sum over data points. Shuffled mini-batches of the training data can then be used to create an unbiased estimate of gradients to be used in an optimization routine. This allows further speedup over the SGP in log-likelihood calculations by constructing much smaller training data matrices $X_b = [x_1, ..., x_b]^T$ where $b$ is the mini-batch size with $b << n$.

Finally, Ref.~\citenum{deep_kernel_wilson} introduces the concept of deep kernel learning, where a deep neural network (DNN), notated as a function $g$, non-linearly transforms the inputs before feeding them into the kernel evaluation.
\begin{align*}
    \Sigma(X, X) &= [k(g(x_i; \theta_{\mathrm{DNN}}), g(x_j; \theta_{\mathrm{DNN}}); \theta)]_{1 \leq i, j \leq n}.
\end{align*}
The hyperparameters of the model now include $\theta_{\mathrm{DNN}}$, the weights of the DNN, in addition to the kernel hyperparameters $\theta$. The GP kernel is evaluated over training data points in a latent space representation constructed by the DNN as opposed to the space of the data itself. The value of this approach is to capture any non-stationary or hierarchical behavior in the training data which typical GPs with stationary kernels cannot model. We pair this concept with the SGP and SVGP to create the deep kernel sparse variational GP (DKSGP) and the deep kernel stochastic variational GP (DKSVGP) with the goal to obtain increased predictive accuracy on the training data while maintaining the speedup that the SGP/SVGP offers. The weights of the DNN, and thus the exact non-linear mapping used, are learned jointly during optimization of the kernel hyperparameters and inducing points locations/mean/covariance by maximizing the log-likelihood of the observations or the ELBO. Additional penalty terms can be added to regularize the weights of the network. 

\subsection{Creating and Optimizing Gaussian Processes in Python}
\label{sec:gpytorch}

We use the Python package \texttt{GPyTorch} for constructing the approximate GP models.\cite{gardner2018gpytorch} \texttt{GPyTorch} provides a modular interface to define both exact and approximate GPs in terms of its base components: a mean function, a kernel function, inducing point approximations, and prediction strategies and distributions that can be used with either exact (in the case of the SGP and DKSGP) or variational inference (in the case of the SVGP and DKSVGP). \texttt{GPyTorch} integrates naturally with \texttt{PyTorch}, a Python package that provides an interface for operating on matrices and tensors with built-in auto-differentiation.\cite{paszke2017automatic} The \texttt{PyTorch} package can be used for creation and training of deep neural networks as well as for other differentiable models. The \texttt{PyTorch} auto-differentiation feature and set of built-in optimizers and used in this work for the training of GP models and deep neural networks in the case of the DKSGP and DKSVGP. In addition, \texttt{GPyTorch} and \texttt{PyTorch} integrate with the \texttt{Pyro} Python package, a probabilistic modelling framework with a focus on variational inference.\cite{bingham2019pyro} \texttt{Pyro} also provides inference utilities through Markov chain Monte Carlo (MCMC), which we utilize for calibration of the UNEDF1 parameters using the approximate GP emulators discussed in this work.

\section{EXPERIMENTAL DESIGN}
\label{sec:experiment}

In this section, we describe the UNEDF1 simulations (Sec.~\ref{sec:unedf1_sim}), our data pre-processing steps (Sec.~\ref{sec:preprocess}), our experiments to understand the performance the approximate emulators, enumerating the choices we made during modeling (Sec.~\ref{sec:approx_emulators}), and finally how we constructed a comparison emulator based on prior work (Sec.~\ref{sec:multivariate}).

\subsection{The UNEDF1 Simulations}
\label{sec:unedf1_sim}

Reference~\citenum{schunck2020calibration} provides extended background on the set of UNEDF1 simulations used in this work. Briefly, the UNEDF1 simulation was run with 500 distinct sets of the 12 UNEDF1 parameter values that fill out a volume of the 12-dimensional parameter space. These parameter values were selected using maximin, space-filling Latin hypercube sampling. Bounds on this volume were chosen to encapsulate the 95\% confidence intervals on the calibrated values of the parameters of the UNEDF0, UENDF1, and UNEDF2 simulators found in prior work.\cite{PhysRevC.82.024313, PhysRevC.85.024304, PhysRevC.87.044320}. For each of these $500$ parameter combinations, the binding energy was evaluated for UNEDF1 for 75 nuclides. These nuclides were balances between light and heavy nuclei and concentrated near magic number nuclei in Z and N. These data have been used for Bayesian calibration with systematic discrepancy in \citenum{higdon2015bayesian,mcdonnell2015uncertainty,schunck2020calibration}.

\subsection{Data Pre-processing}
\label{sec:preprocess}

In this work, we treat proton number $Z$ and neutron number $N$ as inputs to the emulator in addition to the $12$ UNEDF1 simulation parameters. We first take the $500$ simulation parameter sets, the simulation binding energies predictions, and $75$ (Z, N) pairs to create input and output data matrices for use in training and evaluating the emulator. We store the simulation parameters as a matrix with $500$ rows and $12$ columns then stack it $75$ times, enumerating over and appending as additional columns $Z$ and $N$, producing a data matrix $X \in \mathbb{R}^{n \times d}$ with $n = 37,500$ rows and $d = 14$ columns. The simulation-predicted binding energies are similarly placed in a vector $Y \in \mathbb{R}^{n}$. Next, we removed rows corresponding to simulation parameter sets and $(Z, N)$ combinations that either produced error codes during the DFT simulation or produced non-physical predictions of the nuclei properties. This step left $n = 37,116$ rows. We scaled the input data matrix $X$ so that each column varied from $[-1, 1]$ and scaled the target binding energies $Y$ by subtracting their mean and scaling by their standard deviation. The test data, the measured binding energies of the $75$ nuclides were scaled using the same mean and standard deviation. This pre-processing step places the data on a normalized scale that makes it easier to set priors on kernel hyperparameters. Binding energy predictions can be transformed back to their original scale (in units of MeV) by reversing this scaling process. The input points and corresponding target values were shuffled and 80\% of them were chosen for training while 20\% were kept as a validation set. The number of input points used for training was $n_{\mathrm{train}} = 29,693$ and for validation was $n_{\mathrm{val}} = 7,423$, forming two input data matrices $X_{\mathrm{train}} \in \mathbb{R}^{n_{\mathrm{train}} \times d}$ and $X_{\mathrm{val}} \in \mathbb{R}^{n_{\mathrm{val}} \times d}$ as well as two target vectors $Y_{\mathrm{train}} \in \mathbb{R}^{n_{\mathrm{train}}}$ and $Y_{\mathrm{val}}\in \mathbb{R}^{n_{\mathrm{val}}}$. The test data--the observed binding energies of the 75 nuclides--form a differently shaped input matrix $X_{\mathrm{test}} \in \mathbb{R}^{75 \times 2}$, composed of only the scaled $Z$ and $N$ values for the observed nuclides, and target vector $Y_{\mathrm{test}} \in \mathbb{R}^{75}$. When performing inference to find simulation parameters that best fit these data, a $12$-dimensional vector is chosen from the scaled input space and concatenated to each row of $X_{\mathrm{test}}$ to form a matrix of the same size as the training and validation data, allowing the GP emulator to make binding energy predictions for the 75 nuclides.

\subsection{Parameterizing Approximate Emulators}
\label{sec:approx_emulators}

We developed a set of experiments to evaluate the performance of the approximate GPs. For all four types of approximate GP, we vary the number of inducing points used between 2 and 506, spaced by 8, producing 64 different emulators for each emulator type. For the DKSGP and DKSVGP, we additionally use a DNN architecture that maps the the 14 dimensional input to $2$ dimensions using $3$ hidden layers of dimension $100$, $50$, and $5$. The Rectified Linear Unit (ReLU) activation function is applied to the output of each hidden layer. The output of the final layer is scaled to bound the values of the training data in the $2$-dimensional latent space between $-1$ and $1$. This architecture is a modified version of that used in Ref.~\citenum{deep_kernel_wilson}, dividing the dimensionality of the hidden layers by 10 and reducing the number of latent dimensions from $4$ to $2$. We found that reducing the model complexity allowed the emulator to make faster predictions without a loss of expressiveness or prediction accuracy on the validation set. 

The Guassian process is additionally parameterized by its kernel. For each emulator, we used an scaled squared exponential kernel of the form
\begin{align*}
    k(x, x') &= \sigma_s \exp\left(- \frac{1}{2} \sum_{i = 1}^{d} \left(\frac{x_i - x_i'}{l_i}\right)^2 \right).
\end{align*}
This kernel models covariance as exponentially decaying with respect to the scaled distance between points in the input space. This kernel has several hyperparameters: a scale $\sigma_s$ which is the process variance -- the value to which predictive variance converges as distance decreases -- and $d$ length scale parameters $\{l_i\}_{1 \leq i \leq d}$, which scale the distance between locations in each dimension of the input data. 

Finally, the problem of regression with the GP is parameterized by a noise term, $\sigma$, incorporated in the evaluations of the likelihood of observations under the GP prior and predictive distributions. This noise term is essential for modelling of real-world data, where observations are corrupted by noise, and for stochastic simulations, but may not be expected to be necessary for non-stochastic simulations like UNEDF1. We use the noise term here to protect against the effects of numerical instabilities during matrix operations such as decomposition or inversion. The noise term enters into the diagonal of a matrix which is inverted during prediction, acting as regularization to guarantee matrix invertibility\cite{andrianakis2012effect}. We bound the noise term between $10^{-8}$ and $1$ and allow the exact value to be determined through training of the emulator.

\subsection{Constructing a Multivariate Emulator Comparison}
\label{sec:multivariate}

We constructed a version of the multivariate emulator as used in Refs.~\citenum{emulation_higdon_2008,higdon2015bayesian,schunck2020calibration} to act as a comparison for emulator performance. This emulator avoids treating the proton number $Z$ and neutron number $N$ as emulator inputs by treating the binding energy predictions of the 75 nuclides as multivariate outputs. The data are arranged so that $X \in \mathbb{R}^{n\times d}$ where $n = 500$ for the $500$ UNEDF1 parameter sets and $d = 12$ for the 12 UNEDF1 parameters. Additionally, the target binding energies are arranged so $Y \in \mathbb{R}^{n \times p}$ where $p = 75$ for the $75$ nuclides. To construct the multivariate emulator, the binding energy matrix $Y$ has its mean subtracted and is scaled by its standard deviation. It is then decomposed using Principal Component Analysis (PCA), decomposing the matrix $Y$ into a set of basis vectors $K$ and a set of weights $W$:
\begin{align*}
    Y &= W K^T
\end{align*}
where $W \in \mathbb{R}^{n \times d}$ and $K \in \mathbb{R}^{p \times d}$. Next, for each basis vector $j$ in $K$, a GP is used to predict a weight--an entry in $W$--for each element in $X$:
\begin{align*}
    W_{\cdot, j} &= GP_j(X) \sim \mathcal{N}\left(\mu(X), \Sigma(X, X; \theta) \right)
\end{align*}
where the notation $W_{\cdot, j}$ indicates that the $j^{\mathrm{th}}$ column of $W$ is being predicted. An exact GP is trained, meaning the kernel hyperparameters $\theta$ are chosen, to maximize the likelihood of the observed PCA weights under the GP prior. We use the same squared exponential kernel for this emulator as with the approximate GPs. In order to utilize an identical validation set between the approximate and multivariate GP emulators, we fill both $X$ and $Y$ with \texttt{NaN} values and use an implementation of probabilistic PCA to construct the basis and weight matrices with missing data\cite{tipping1999probabilistic}\footnote{For implementation, see \url{https://github.com/shergreen/pyppca}.} At new inputs $x^* \in \mathbb{R}^{d}$, the predictive distribution over target values $y^* \in \mathbb{R}^{p}$ becomes a modified version of Eqn.~\ref{eqn:pred} that scales the predictive mean and covariance matrix by the basis vectors in $K$:
\begin{align*}
    y^* \sim \mathcal{N} \left(\tilde{\mu} K^T, K^T \tilde{\Sigma} K \right)
\end{align*}
where $\tilde{\mu} \in \mathbb{R}^{d}$ and $\tilde{\Sigma} \in \mathbb{R}^{d \times d}$ are the predictive mean and covariance matrix on the weights, constructed through evaluation of the $d$ GP models. $\tilde{\Sigma}$ is diagonal, containing the predictive variances of the weights from the GP models. 

\section{TRAINING}
\label{sec:training}

We performed joint optimization of the kernel hyperparameters, the likelihood noise term, the inducing point locations (in the case of the SGP and DKSGP), the inducing point mean/covariance (in the case of the SVGP and DKSVG), and DNN weights (in the case of the DKSGP and DKSVGP) to find parameter values that maximized the log-likelihood of observations from the training data set under the GP prior. For the SGP and DKSP Eqn.~\ref{eqn:L_SGP} was maximized while the ELBO from Ref.~\citenum{hensman2013gaussian} was maximized for the SVGP and DKSVGP.

We used a different optimizer and optimization scheme for each emulator type. For the SGP, we used the Limited-memory Broyden–Fletcher–Goldfarb–Shanno (LBFGS) optimization algorithm implemented in \texttt{PyTorch}.\cite{lbfgs} We used a learning rate of $0.01$ for all SGP emulators. For the SVGP, we used two optimizers: one for the variational parameters, namely the inducing point mean and covariance matrix, and one for the additional kernel and likelihood hyperparameters. The inducing point mean and covariance matrix were learned in their ``natural'' representation, using the \texttt{TrilNaturalVariationalDistribution} in \texttt{GPyTorch}, and were updated using natural gradient descent implemented in \texttt{GPyTorch}.\cite{hensman2013gaussian, salimbeni2018natural} We used a learning rate of $0.1$ for the natural gradient descent optimizer. The kernel and likelihood hyperparameters were learned using the Adam optimizer with a learning rate of $0.01$.\cite{adam} The Adam optimizer was used with a learning rate of 0.01 for the DKSGP with a ``weight decay'' of $10^{-4}$, which penalizes values of the DNN weights through an L2 loss. For the DKSVGP, we used the Adam optimizer with a learning rate of 0.01 for the kernel and likelihood hyperparameters and the natural gradient descent optimizer with a learning rate of 0.1 for the inducing point mean and covariance matrix. Additionally, a weight decay of $10^{-4}$ was used to regularize the DNN weights. The Adam optimizer was chosen for use with the SVGP, DKSGP, and DKSVGP since it supports parameter updates with mini-batches and provides faster optimization steps when optimizing the many more parameters from the DNN in the DKSGP and DKSVGP emulators.

Each emulator was trained for 4000 epochs, where an epoch is defined as one traversal of the training data. The SVGP and DKSVGP emulators were trained using mini-batches of the training set of size $n_b = 512$, viewing only a small portion of the training data at each optimization step. Mini-batches were shuffled between each epoch to ensure computed gradients with respect to the negative log marginal likelihood were unbiased. 

\begin{figure}[h]
    \centering
    \includegraphics[width=\textwidth]{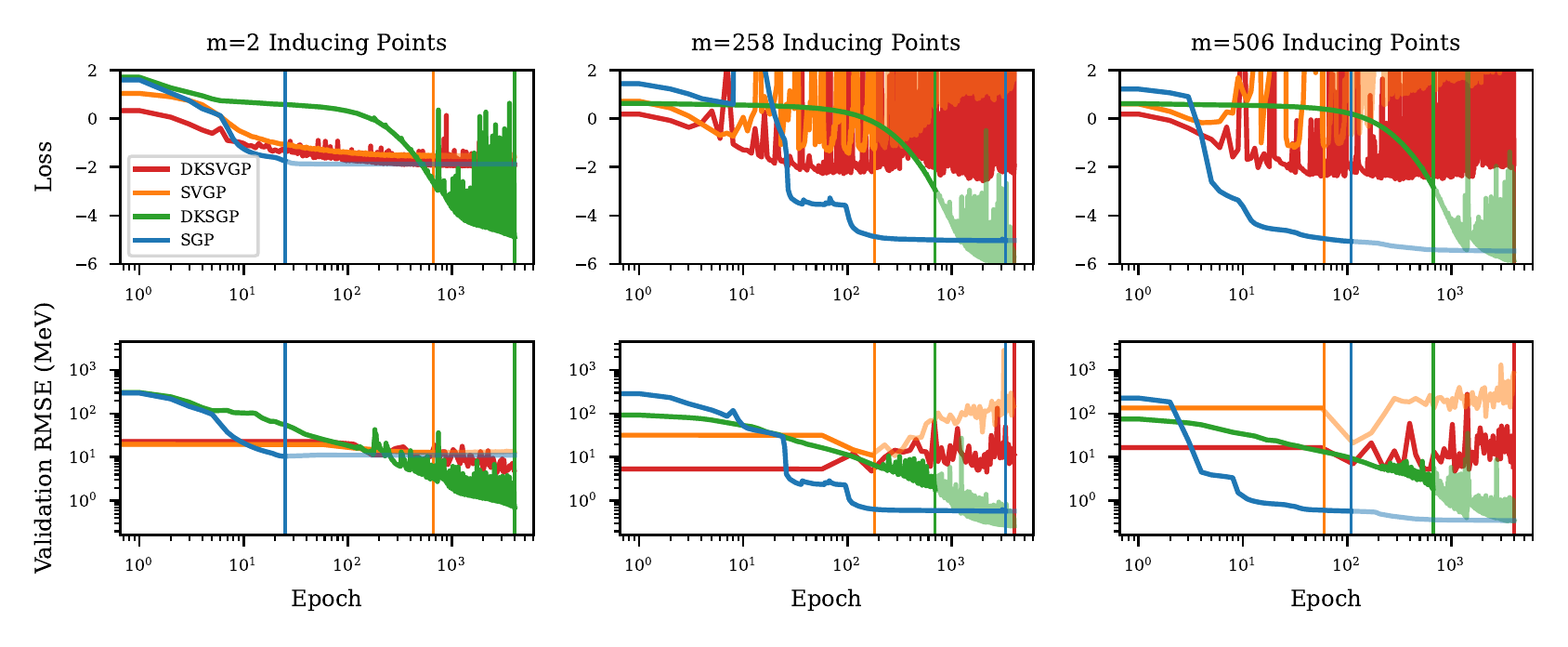}
    \caption{The value of the loss function (top) and validation set RMSE (bottom) while optimizing the hyperparameters of the emulators during training for emulators with 2 (left), 258 (middle), and 506 (right) inducing points. The loss function is the negative marginal log-likelihood or its upper bound on the training data. The intensity of line shading and the colored vertical lines indicate at which epoch the training produced the last best-performing (lowest RMSE on the validation set) and numerically stable (on the test set) model.}
    \label{fig:training}
\end{figure}

Convergence was evaluated through visual inspection of the loss function value over the training epochs. Figure~\ref{fig:training} shows example traces of the loss function--the negative log-likelihood of the observations under the GP prior--and validation set prediction accuracy in terms of RMSE through the 4000 epochs of training for the different emulators. We observed convergence in the loss function after 4000 epochs for the SGP, SVGP, and DKSVGP emulators. The DKSGP emulators typically showed potential for further accuracy gains on the training and validation data sets with continued training. 

Emulators typically showed instability during their training which made training for longer than 4000 epochs an unnecessary endeavor. During training, the loss function could ``spike'' before returning to a normal level, indicating that there were regions of parameter space that the optimizer had trouble with using a uniform learning rate. Future work should consider experimenting with learning rates other than those used here to investigate this effect further. Typically, the optimization could recover from these spikes, bringing the loss function back to typical values and updating closer towards a minimum. Sometimes however, the emulator would produce a covariance matrix with \texttt{NaN} values, indicating that the hyperparameter values chosen in an optimization step produced invalid values when evaluating the kernel. Additionally, the emulators could train without signs of instability, but produce invalid covariance matrices, either with \texttt{NaN} values or with non-positive-definite or non-symmetric properties, on unseen data, namely the validation data set or at random points in the parameter space. Given this unstable behavior during training, we opted to save two versions of each model: the model after 4000 epochs and the model that both produced the lowest RMSE prediction error on the validation set and produced covariance matrices with the correct properties at 10 random locations of the input data set. This latter model is the model which will be useful in application, while the former shows the potential for the emulator's use in prediction on the test data alone as well as its potential as an emulator had stability not been an issue. 
We additionally observed that training of the emulators could get stuck in a local minimum, particularly for the SGP models. This indicates that gradient-based optimization routines may not be sufficient for training inducing point based approximate GPs and the global minimum in the loss function may need to be determined through e.g. multiple training runs or optimization methods such as annealing.

\FloatBarrier
\section{RESULTS}
\label{sec:results}

We evaluated the performance of the trained emulators in three contexts: in their ability to accurately predict the validation set and the runtime during prediction (Sec.~\ref{sec:evaluation}) and the result and stability of finding maximum likelihood estimates (Sec.~\ref{sec:optimize}) and posterior distributions (Sec.~\ref{sec:calibration}) of the 12 UNEDF1 parameters with respect to the measured binding energies of the 75 nuclides.

\subsection{Accuracy and Timing}
\label{sec:evaluation}

We evaluated the accuracy and prediction time of the set of trained emulators on the validation data set. We call this process ``evaluation.'' Accuracy is measured as the root mean squared error (RMSE) between the predictive mean of the GP and simulation-predicted binding energies from the validation data set. The RMSE is defined as:
\begin{align*}
    RMSE &= \sqrt{\frac{1}{\vert X_\mathrm{val} \vert} \sum_{j = 1}^{\vert X_\mathrm{val} \vert} \left(y_{j}^{\mathrm{sim}} - y_{j}^{\mathrm{emu}} \right)^2}
\end{align*}
where $y_j^{\mathrm{sim}}$ is the UNEDF1 simulation predicted binding energy and $y_j^{\mathrm{emu}}$ is the emulator predicted binding energy of the $j^{\mathrm{th}}$ element in the validation set.
Additionally, we measured the single-threaded runtime of calculating the RMSE of a single input point in the validation data set. We varied the input point $30$ times to produce a set of samples of the prediction time. We report the mean and standard deviation of the prediction time samples as the typical prediction time of the emulator. 

\begin{figure}[h]
    \centering
    \includegraphics[width=\textwidth]{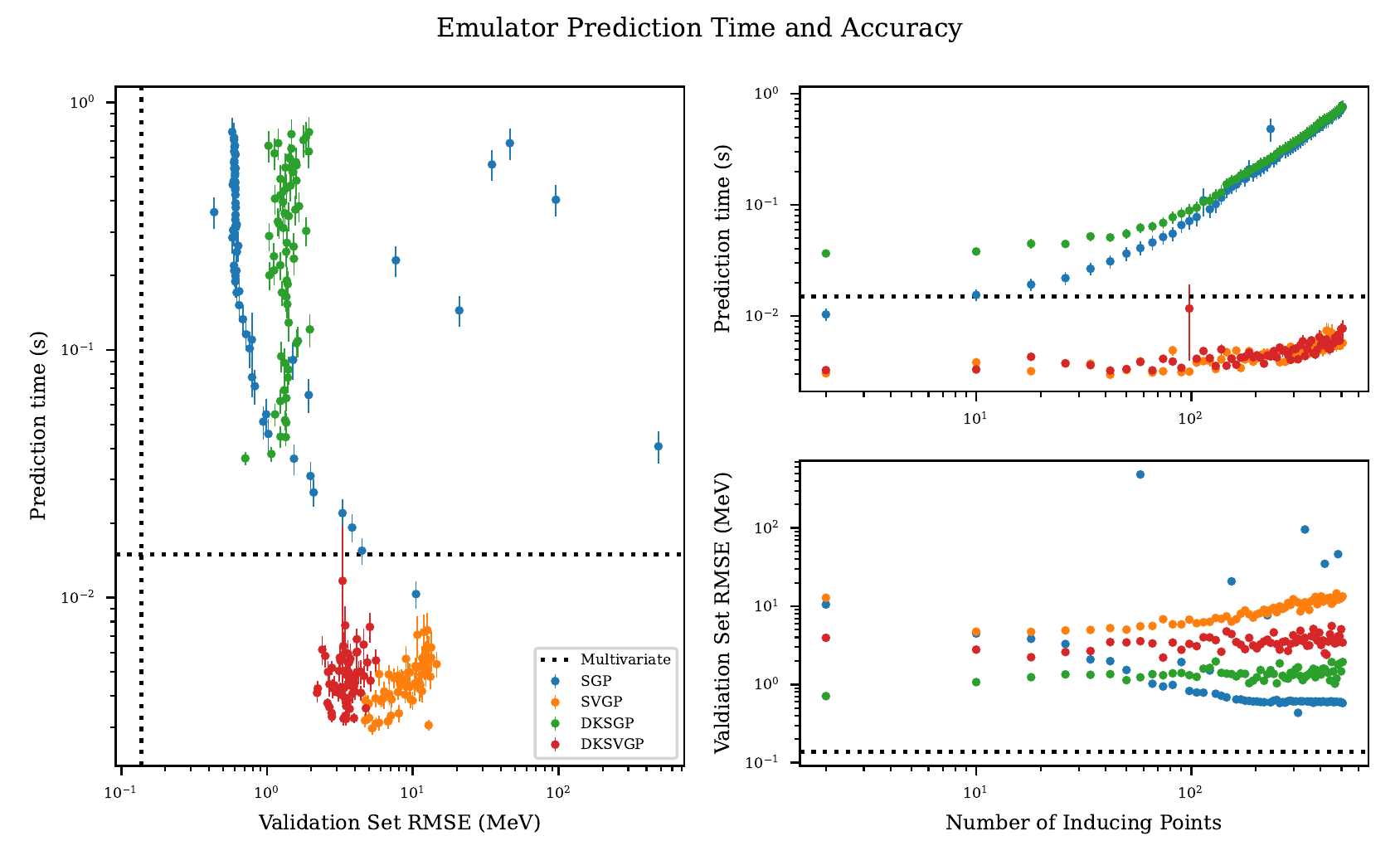}
    \caption{The RMSE of binding energy predictions on the validation set and single-threaded single-point prediction runtimes for the set of trained emulators that provided the best validation set RMSE during training and produce numerically stable predictions within the parameter space. Points are colored by the emulator type, and dashed lines show the performance of the multivariate emulator on the same metrics. The left panel visualizes the trade-off in accuracy and runtime, while the panels on the right visualize the prediction accuracy and runtime with respect to the number of inducing points used.}
    \label{fig:accuracy_and_runtime}
\end{figure}

\begin{figure}[h]
    \centering
    \includegraphics[width=\textwidth]{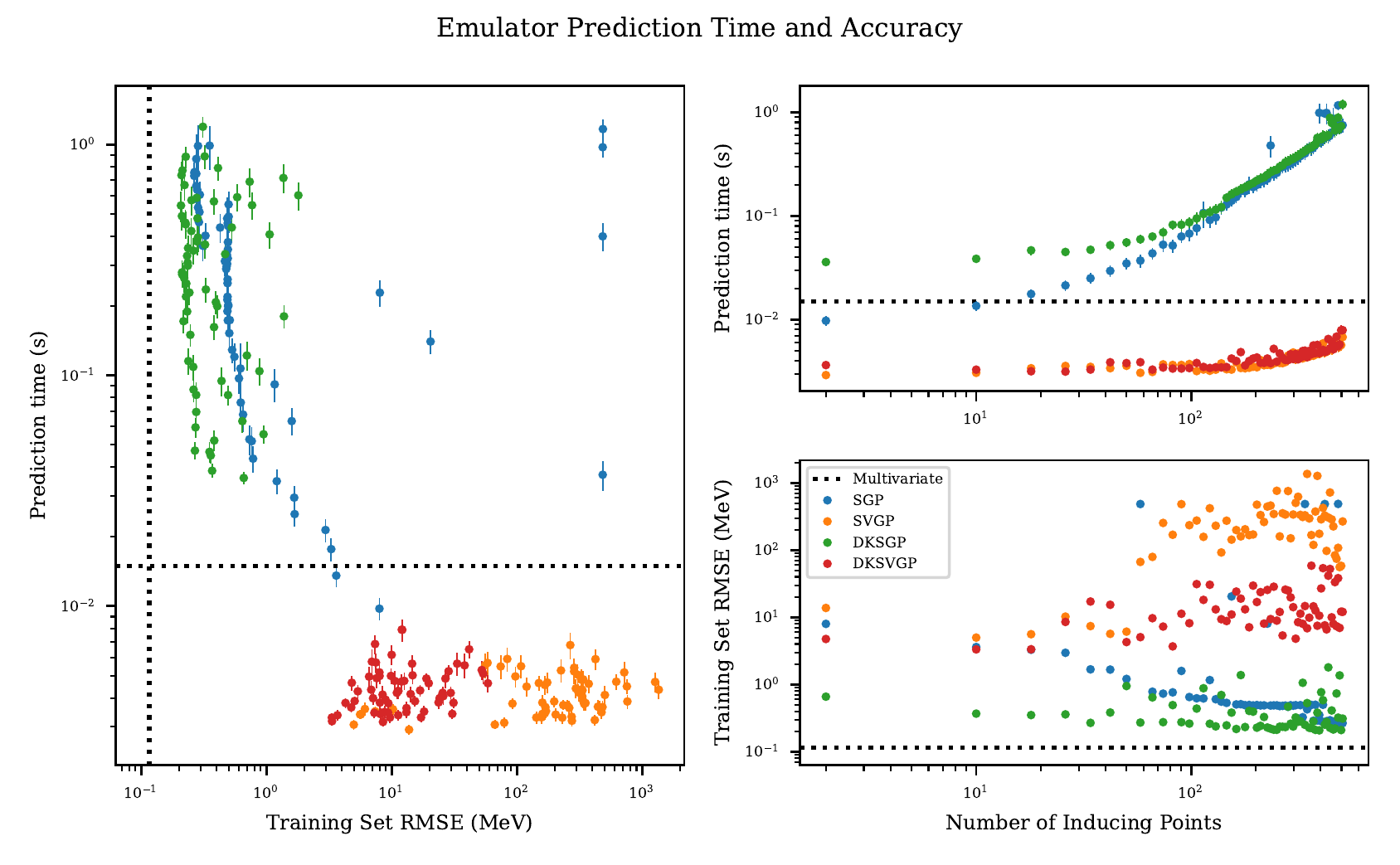}
    \caption{The RMSE of binding energy predictions on the training data set and single-threaded single-point prediction runtimes for the set of fully trained emulators. Points are colored by the emulator type, and dashed lines show the performance of the multivariate emulator on the same metrics. The left panel visualizes the trade-off in accuracy and runtime, while the panels on the right visualize the prediction accuracy and runtime with respect to the number of inducing points used.}
    \label{fig:accuracy_and_runtime_train}
\end{figure}

Figure \ref{fig:accuracy_and_runtime} shows the accuracy and prediction time for all of the trained emulators that achieved the best validation set RMSE during training and also provided stable predictions in the parameter space. Figure \ref{fig:accuracy_and_runtime_train} visualizes the same for the fully trained emulators instead computing the RMSE on the training data, since validation set predictions may be unstable. Outliers in accuracy measurements in both figures for the SGP emulators are due to the aforementioned issues of stability during training and training converging to a local instead of global minimum in the loss function. Table~\ref{tab:best_emulators} summarizes the accuracy and runtime of the best performing emulators in each category. The emulators that best predict the validation set are the SGP, followed by the DKSGP, the DKSVGP, and the SVGP, while none of the approximate GP emulators surpassed the multivariate emulator in prediction accuracy. In contrast, and as Fig.~\ref{fig:accuracy_and_runtime_train} shows, the fully trained DKSGP emulators provided the best prediction accuracy on the training data set, approaching the prediction accuracy of the multivariate emulator. This indicates there are features of the data--non-stationary or hierarchical structure--that cannot be captured with the SGP but can be with the DKSGP. While the SVGP and DKSVGP performed poorly in prediction accuracy, their performance in prediction runtime was the best among all emulators, achieving an order of magnitude faster prediction time than most others, and additionally were faster than the multivariate emulator.

\begingroup
\setlength{\tabcolsep}{10pt} % Default value: 6pt
\renewcommand{\arraystretch}{1.5} % Default value: 1
\begin{table}[h]
    \centering
    \begin{tabular}{cccc}
\textbf{GP Type} & \textbf{Number of Inducing Points} & \textbf{Runtime (s)} & \textbf{Accuracy (MeV)} \\ \hline \hline
\multirow{2}{*}{SGP} & 314 & $0.360 \pm 0.279$ & \textbf{0.434} \\ \cline{2-4}
	 & 2 & $ \mathbf{0.0103 \pm 0.0069}$ & 10.5 \\ \cline{2-4} \hline
\multirow{2}{*}{SVGP} & 18 & $0.00319 \pm 0.00104$ & \textbf{4.70} \\ \cline{2-4}
	 & 42 & $ \mathbf{0.00297 \pm 0.00092}$ & 5.25 \\ \cline{2-4} \hline
DKSGP & 2 & $ \mathbf{0.0365 \pm 0.0116}$ & \textbf{0.709} \\ \hline
\multirow{2}{*}{DKSVGP} & 74 & $0.00412 \pm 0.00183$ & \textbf{2.20} \\ \cline{2-4}
	 & 42 & $ \mathbf{0.00323 \pm 0.00102}$ & 3.48 \\ \cline{2-4} \hline
Multivariate & & $ \mathbf{0.015 \pm 0.003}$& \textbf{0.138} \\ \hline    
    
    \end{tabular}
    \caption{A summary of the best performing emulators for the different GP types in terms of prediction accuracy (RMSE) and prediction runtime. Runtime is reported as the mean and standard deviation of prediction times among 30 samples. Best-in-class performance metrics have bold face type in the table for visual clarity.}
    \label{tab:best_emulators}
\end{table}
\endgroup

\FloatBarrier
\subsection{Optimization}
\label{sec:optimize}

While Sec.~\ref{sec:evaluation} evaluates the performance of the emulators when predicting simulator results, it is also valuable to understand their performance as utilities for inference; that is, understanding for which simulation parameters the emulator (and hence simulator) provide predictions that best match a set of observed binding energies. 

\begin{figure}[h]
    \centering
    \includegraphics[width=\textwidth]{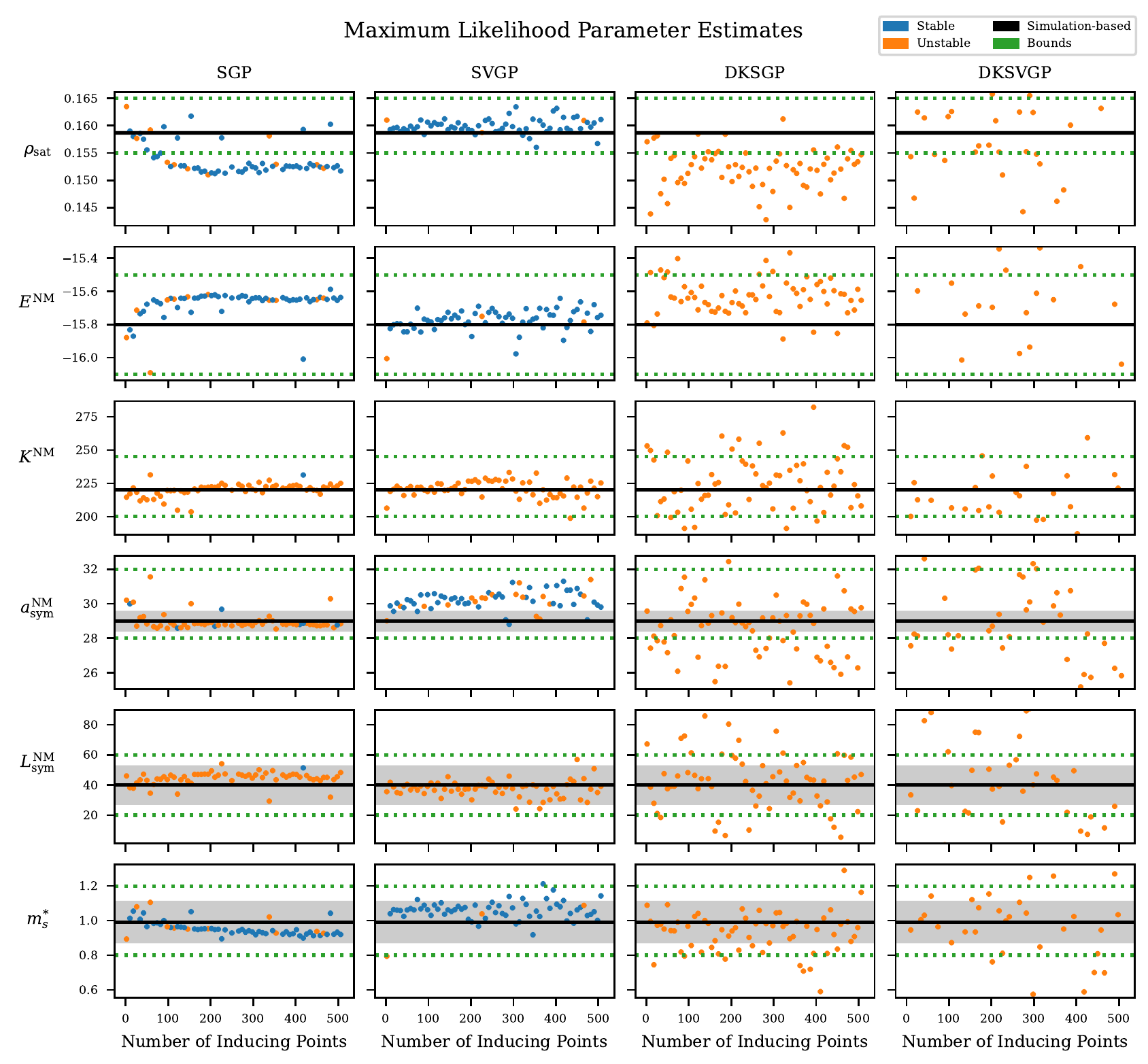}
    \caption{The estimates for each of the UNEDF1 simulation parameters that maximize the likelihood of the observed binding energies under the emulator's predictive distribution. These estimates are found through optimization with LBFGS. Each panel shows the parameter value at maximum likelihood for each emulator type as a function of the number of inducing points. Points are colored by whether the maximum likelihood estimate was stable over 10 independent rounds of optimization with LBFGS. Missing points lie outside of the plotting region. Each panel shows the maximum likelihood estimate of the parameter values (black line) and a derived $1\sigma$ uncertainty range (gray rectangle) when optimization was performed using the simulation directly in Ref.~\citenum{schunck2020calibration}. Green dotted lines show the bounds of the parameter values in the training set. This figure is continued in Fig.~\ref{fig:optimization2}.}
    \label{fig:optimization1}
\end{figure}

\begin{figure}[h]
    \centering
    \includegraphics[width=\textwidth]{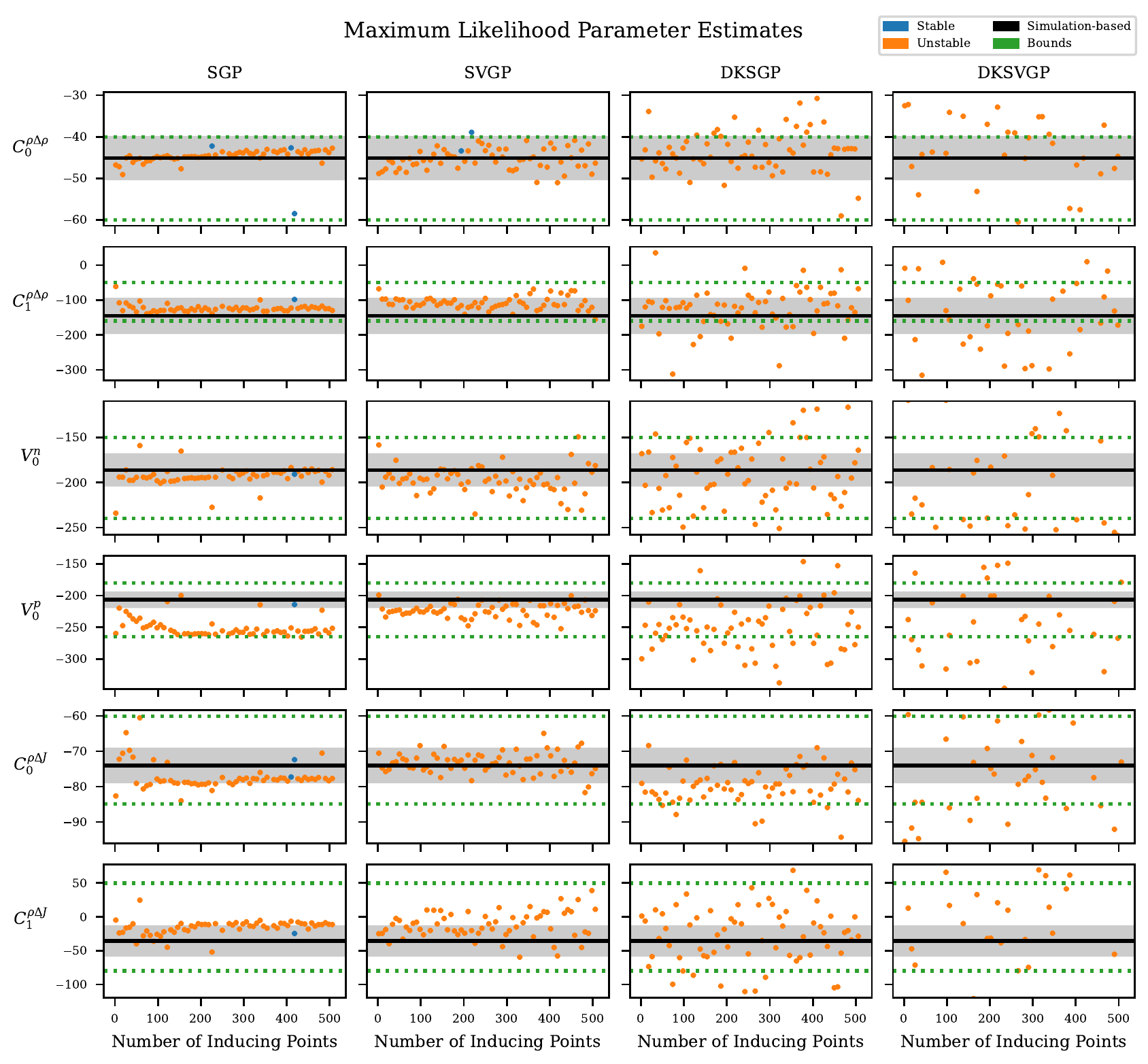}
    \caption{A continuation of Fig.~\ref{fig:optimization1}}
    \label{fig:optimization2}
\end{figure}

To understand this, we used the LBFGS optimizer to find estimates of the simulation parameters that maximize the likelihood of the observed binding energies for the 75 nuclides under the emulator predictive distribution. To evaluate whether the emulators admitted only one or several optima in the likelihood, we performed optimization of the likelihood in 10 rounds, each time starting at a different randomly chosen point in the parameter space and iterating the LBFGS optimizer for 100 steps. To evaluate whether the optimization converged to the same location over the 10 rounds, we computed the per-parameter standard deviation of the location of the maximum. If the standard deviation for a parameter was greater than $10^{-4}$, we label that emulator as providing unstable ML estimates for that parameter. Figures \ref{fig:optimization1} and \ref{fig:optimization2} summarize the results of this optimization among the various emulators and compares them to the result of direct optimization using the DFT simulation, using the values from Table 3 in Ref.~\citenum{schunck2020calibration}. 

We observe that the maximum likelihood parameter values are often consistent with the simulation-derived result, with some parameter estimates lying within the error range. Notably, the DKSGP and DKSVGP emulators show greater scatter in the ML estimate as a function of number of inducing points when compared to the SGP or SVGP. This may be due to the warping of the input space on the part of the DNN. Since the DNN weights are retrained for each DKSGP emulator, the warping will be different for each emulator, perhaps also warping the loss surface of the emulator. The SGP emulators exhibit a trend for some parameter estimates with respect to the number of inducing points, ``converging'' towards a consistent estimate as the number of inducing points and therefore the emulator accuracy increases. With increasing inducing points, the parameter estimates for $\rho_{\mathrm{sat}}$, $E_{\mathrm{NM}}$, $m_s^*$, $C_{0}^{\rho\Delta\rho}$, and $V_{0}^{p}$ trend toward a ``convergence point''. Only for $C_{0}^{\rho\Delta\rho}$ do the ML estimates converge to the simulation-derived estimate, whereas for the rest they converge to a point offset from the simulation-derived estimate. The SVGP yields ML estimates broadly consistent with the simulation-derived estimates, but with more scatter and less structure than the SGP. Finally, all emulators except the SVGP produced ML paramater estimates that lie outside the bounds of the input data, in a region where the emulator is extrapolating rather than interpolating output of the simulation, and should be interpreted with caution.

\FloatBarrier
\subsection{Calibration}
\label{sec:calibration}

In addition to maximum likelihood estimates of the simulation parameters values (Sec.~\ref{sec:optimize}) under the emulator's predictive distribution, it is desirable to produce Bayesian posterior distributions over plausible parameter values. We call this process of deriving posterior distributions ``calibration.'' We set a uniform prior over the range $[-1, 1]$ for each of the parameters, which are the bounds of the simulation parameter space volume explored with the UENDF1 simulations. We used Markov chain Monte Carlo (MCMC) with the No U-Turn Sampler (NUTS) implemented in the \texttt{Pyro} Python package to obtain samples from the resulting posterior distribution.\cite{nuts} We set the maximum tree depth size of the sampler to 5 and used 100 ``warmup'' steps to tune the sampler step size to achieve a target acceptance rate of $70\%$. We then obtained $10,000$ samples from the posterior distribution. Convergence of the Markov chain was assessed using the split Gelman-Rubin $\hat{R}$ statistic, computed using \texttt{Pyro} diagnostics.\cite{gelman1992inference} The Gelman-Rubin statistic compares the variance of samples within-chain and between-chain across multiple Markov chains, producing a value a value close to $1$ for each parameter when the chains have converged to the same target distribution. We only used one chain, and so the split Gelman-Rubin statistic is used, which computes $\hat{R}$ by splitting chain in half and computing within-chain and between-chain variances using the two halves. We observed values of $\hat{R}$ were typically close to $1$, with a few at $1.2 - 1.3$ and even fewer with values of $2+$ (unconverged). Efficient sampling was a challenge for the NUTS sampler for the more accurate emulators. The number of effective samples, computed using \texttt{Pyro} diagnostics, was $\sim 10^3 - 10^4$ for the SVGP and DKSGP emulators and for the least accurate SGP emulators. The number of effective samples was only $\sim 10-10^2$ for the remaining SGP and DKSGP emulators.

\begin{figure}[h]
    \centering
    \includegraphics[width=\textwidth]{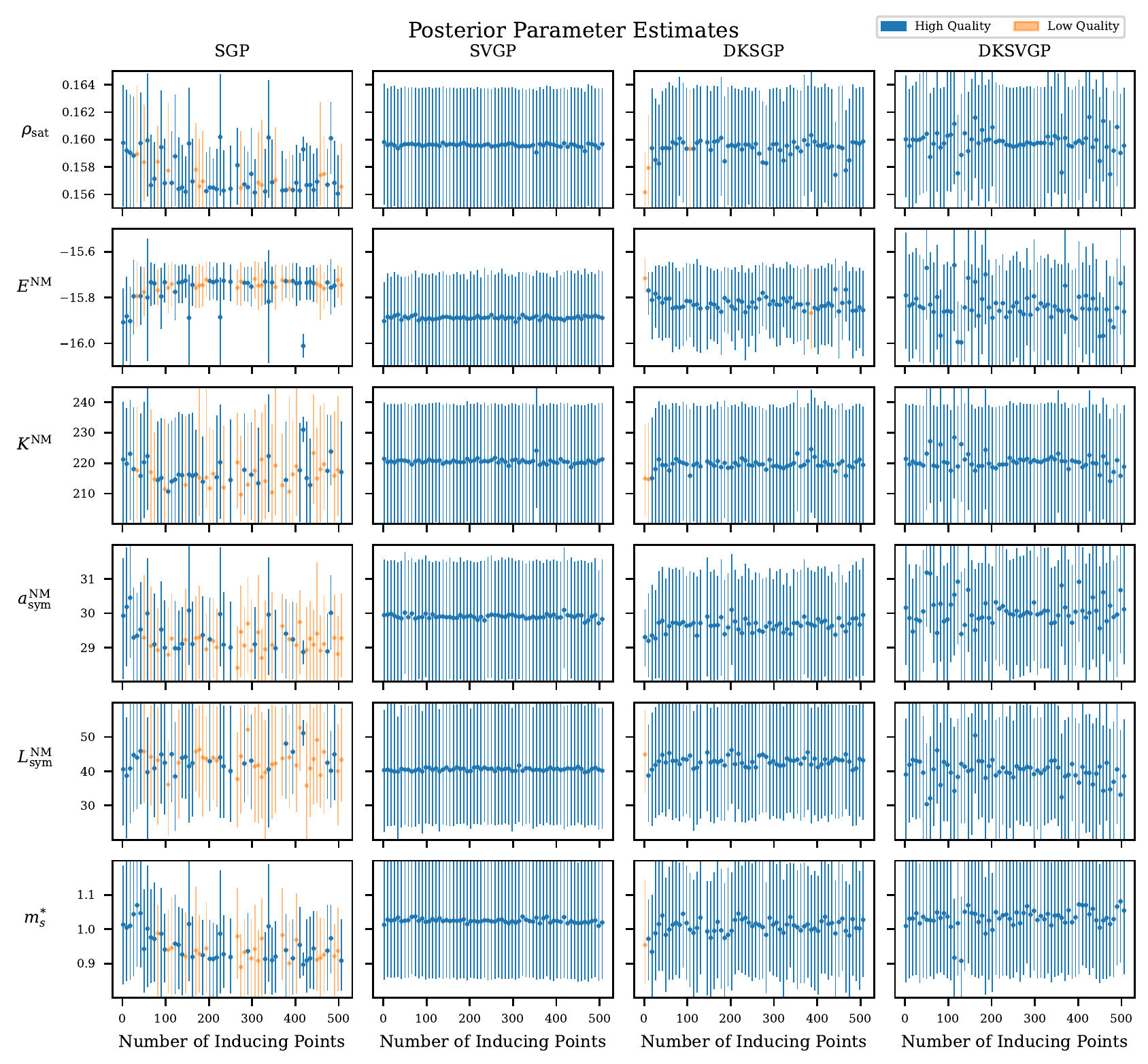}
    \caption{The posterior parameter estimates for each of the trained emulators. The panels visualize the median and 90\% highest posterior density interval of the marginal posterior for each of the UNEDF1 parameters. This figure is continued in Fig.~\ref{fig:calibration_2}.}
    \label{fig:calibration_1}
\end{figure}

\begin{figure}[h]
    \centering
    \includegraphics[width=\textwidth]{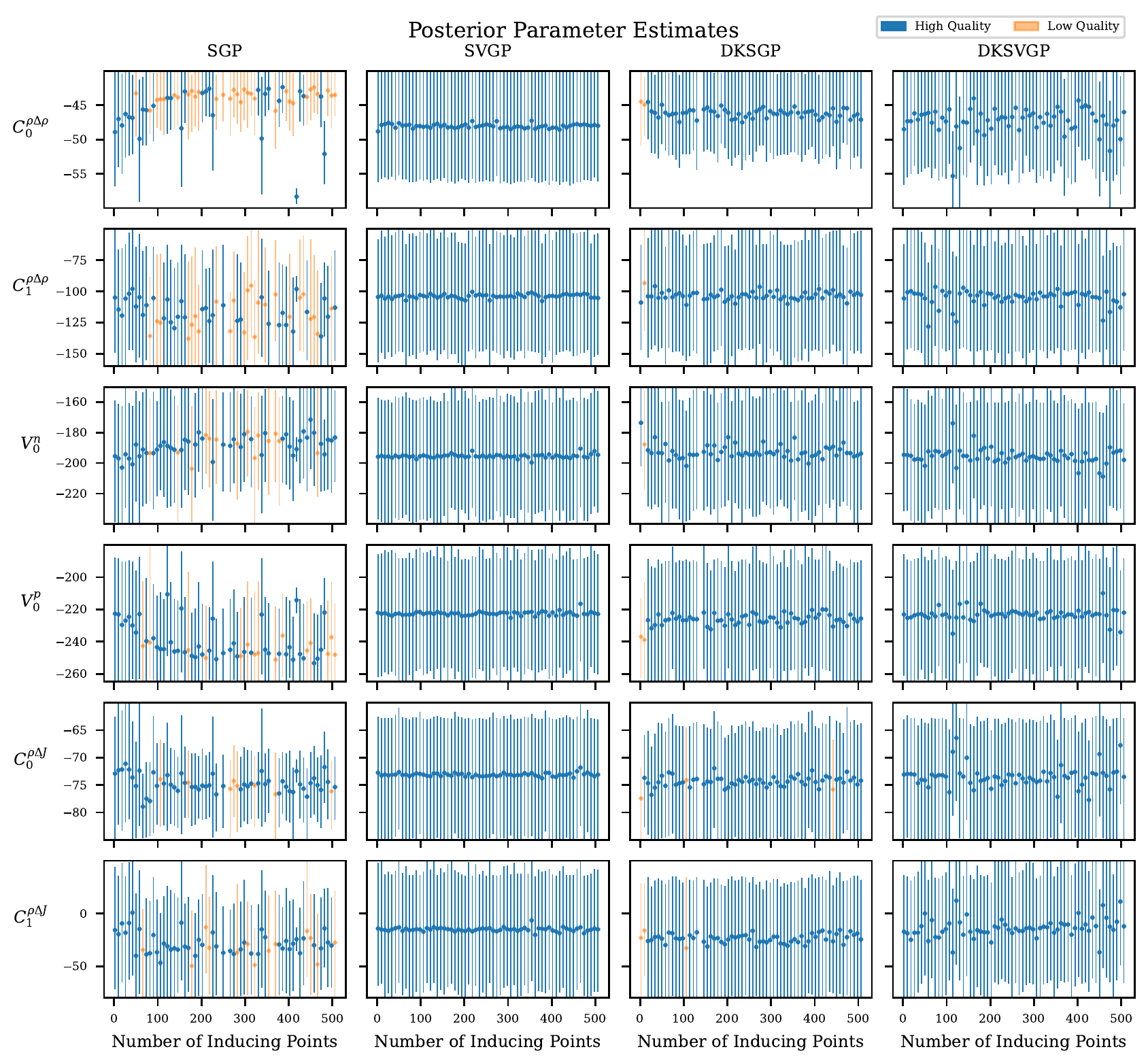}
    \caption{A continuation of Fig.~\ref{fig:calibration_1}.}
    \label{fig:calibration_2}
\end{figure}

Figures \ref{fig:calibration_1} and \ref{fig:calibration_2} show the median and 90\% highest posterior density interval of the marginal posteriors for each UNEDF1 simulation parameter. In these figures, we highlight estimates that are likely to be low quality, that is if the effective sample size of the Markov chain was $<30$ or had $\hat{R} < 1.1$. Table \ref{tab:calibration} provides posterior parameter estimates from the most accurate emulators of each type that had a ``high quality'' Markov chain.

\begingroup
\setlength{\tabcolsep}{10pt} % Default value: 6pt
\renewcommand{\arraystretch}{1.5} % Default value: 1
\begin{table}[h]
    \centering
    \begin{tabular}{c|c|c|c|c}
& \textbf{SGP} & \textbf{SVGP} & \textbf{DKSGP} & \textbf{DKSVGP} \\ \hline\hline  
 Num. Inducing Points & 474 & 18 & 466 & 74 \\ \hline  
 RMSE (MeV) & 0.602 & 4.70 & 1.02 & 2.20 \\ \hline  
$\rho_{\mathrm{sat}}$ & $0.157_{-0.002}^{+0.003}$ & $0.160_{-0.005}^{+0.004}$ & $0.158_{-0.003}^{+0.004}$ & $0.160_{-0.004}^{+0.004}$ \\ \hline 
$E^{\mathrm{NM}}$ & $-15.7_{-0.1}^{+0.1}$ & $-15.9_{-0.2}^{+0.2}$ & $-15.8_{-0.1}^{+0.1}$ & $-15.8_{-0.2}^{+0.2}$ \\ \hline 
$K^{\mathrm{NM}}$ & $217_{- 17}^{+ 18}$ & $221_{- 21}^{+ 19}$ & $216_{- 16}^{+ 21}$ & $219_{- 19}^{+ 20}$ \\ \hline 
$a^{\mathrm{NM}}_{\mathrm{sym}}$ & $28.9_{-0.9}^{+1.1}$ & $300_{-2}^{+2}$ & $29.5_{-1.5}^{+1.4}$ & $29.4_{-1.4}^{+1.7}$ \\ \hline 
$L^{\mathrm{NM}}_{\mathrm{sym}}$ & $42.5_{-10.5}^{+17.3}$ & $40.2_{-16.2}^{+19.6}$ & $450_{-16}^{+15}$ & $46.1_{-15.6}^{+13.8}$ \\ \hline 
$m^{*}_{s}$ & $0.937_{-0.116}^{+0.133}$ & $1.02_{-0.17}^{+0.17}$ & $0.995_{-0.179}^{+0.145}$ & $1.05_{-0.18}^{+0.15}$ \\ \hline 
$C_{0}^{\rho\Delta\rho}$ & $-43.7_{-4.4}^{+3.6}$ & $-47.7_{-7.7}^{+7.7}$ & $-45.4_{-5.1}^{+5.4}$ & $-45.9_{-8.3}^{+5.9}$ \\ \hline 
$C_{1}^{\rho\Delta\rho}$ & $-136_{-  24}^{+  43}$ & $-106_{-  49}^{+  50}$ & $-105_{-  54}^{+  41}$ & $-96.5_{-47.3}^{+46.0}$ \\ \hline 
$V_{0}^{n}$ & $-187_{-  31}^{+  36}$ & $-196_{-  43}^{+  37}$ & $-187_{-  39}^{+  36}$ & $-192_{-  37}^{+  42}$ \\ \hline 
$V_{0}^{p}$ & $-245_{-  20}^{+  28}$ & $-222_{-  37}^{+  39}$ & $-228_{-  37}^{+  35}$ & $-224_{-  40}^{+  35}$ \\ \hline 
$C_{0}^{\rho\Delta J}$ & $-75.9_{-7.0}^{+6.3}$ & $-73.1_{-11.9}^{+10.3}$ & $-74.5_{-9.3}^{+9.0}$ & $-73.5_{-11.4}^{+10.6}$ \\ \hline 
$C_{1}^{\rho\Delta J}$ & $-33.2_{-39.5}^{+48.6}$ & $-15.7_{-63.8}^{+52.6}$ & $-29.7_{-50.1}^{+60.4}$ & $-22.5_{-57.4}^{+56.2}$

    \end{tabular}
    \caption{The median and lower/upper bounds of the 90\% highest posterior density interval of the marginal posteriors for each UNEDF1 simulation parameter. Posterior estimates are reported for each parameter type and for the emulator that had the best prediction accuracy on the validation set and had a Markov chain that was converged ($\hat{R} < 1.1$) and produced $>30$ effective samples for each parameter.}
    \label{tab:calibration}
\end{table}
\endgroup

We observe that for the least accurate emulators, that is the SVGP and DKSVGP emulators, the posterior distributions obtained are largely unconstrained, matching the uniform prior on the most of parameters except $E^{\mathrm{NM}}$ and $C_{0}^{\rho \Delta \rho}$. The posterior distributions for the DKSVGP emulators show more variation with respect to the number of inducing points in comparison to the SVGP. The SGP emulators are accurate enough to allow constraints to be placed on some or all of the simulation parameters. The tightest constraints were placed on the $E^{\mathrm{NM}}$ and $C_{0}^{\rho\Delta\rho}$ parameters, while moderate to loose constrains were placed on the rest. Clear trends of ``convergence'' in the parameter estimate with respect to the number of inducing points is seen with the SGP emulator for the $\rho_{\mathrm{sat}}$, $E^{NM}$, $C_{0}^{\rho\Delta\rho}$, and $V_0^{p}$, similar to the optimization results. This indicates that when using the SGP, more inducing points capture more structure in the data that carry over to accuracy in inference. The DKSGP provided looser constraints across the number of inducing points in comparison to the SGP. This is an effect of the early stopping in training preventing most of the the DKSGP emulators from becoming better predictors of the simulation results. The best performing DKSGP emulator with $m = 2$ inducing points places the tightest constraints on the parameter values. The posterior distributions obtained by this emulator are consistent with the distributions obtained by the best SGP emulators.

\begin{figure}[h]
    \centering
    \includegraphics[width=\textwidth]{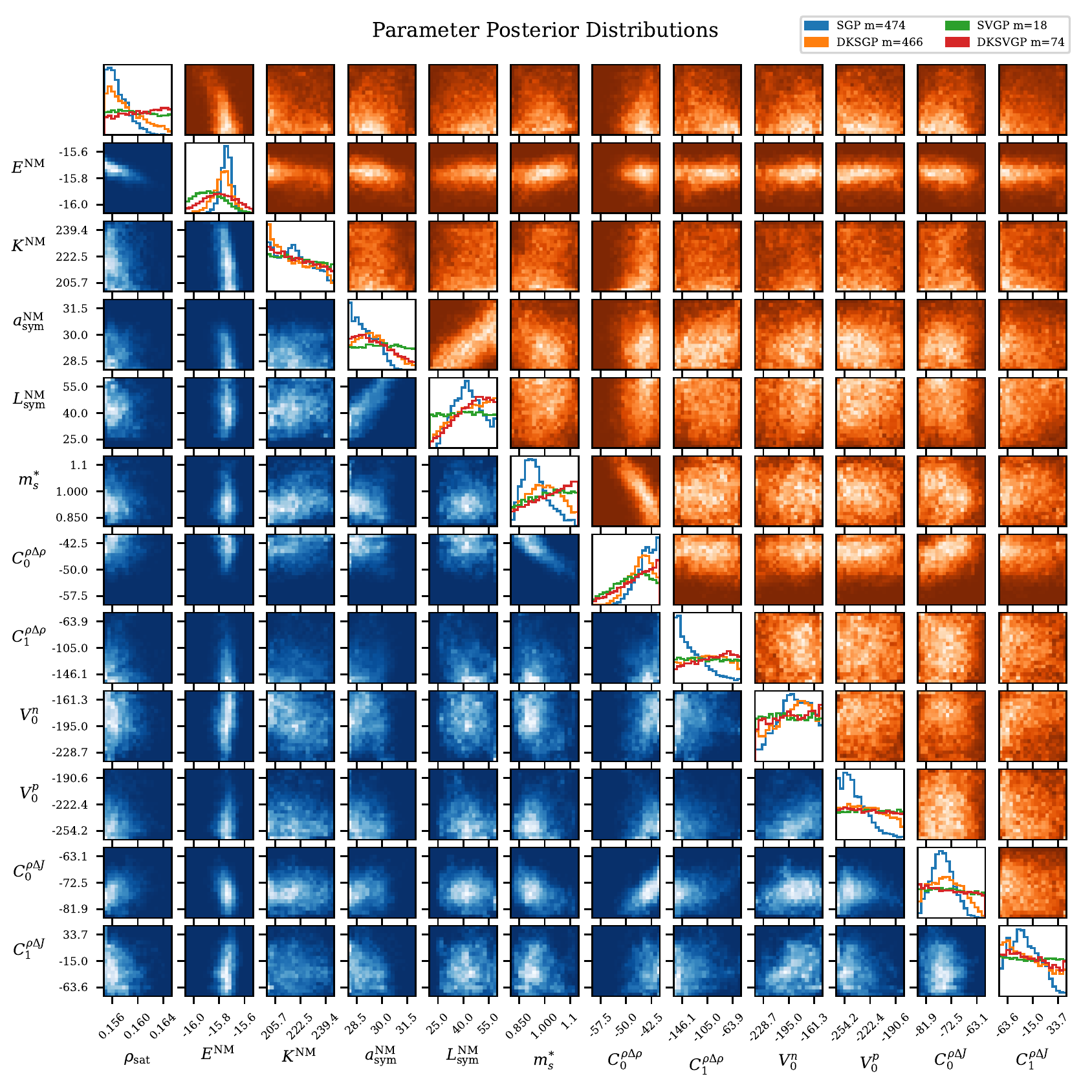}
    \caption{A visualization of 10,000 samples from the posterior distributions of the UNEDF1 parameters for each emulator type. The calibration results visualized are for emulators that had $\hat{R} < 1.1$ and $>30$ effective samples during sampling, and achieved the best RMSE on the validation set. This was the $m = 474$ inducing point SGP, $m = 18$ SVGP, $m = 466$ DKSGP, and the $m = 74$ DKSVGP. Univariate marginal posteriors for all four emulator types are visualized as histograms of the posterior samples on the diagonal. The off-diagonal panels visualize bivariate 2D-histograms of the posterior samples for the SGP (blue) and DKSGP (orange) emulators.}
    \label{fig:posterior}
\end{figure}

To further highlight how the posterior distributions change as the accuracy of the emulator improves, we visualize the univariate and bivariate posteriors of the UNEDF1 simulation parameters derived using the four emulator types in Fig.~\ref{fig:posterior}. The emulators chosen match those from Table~\ref{tab:calibration}, that is they were the most accurate emulators that achieved $\hat{R} < 1.1$ and had $>30$ effective samples during sampling. The marginal posteriors clearly get tighter as the emulator becomes more accurate: from the SVGP, to DKSVGP, to DKSGP, to SGP. Notably, the SVGP and DKSVGP provide moderate constraints for some parameters before they do for others, indicating that some parameter constraints are less sensitive to the accuracy of the emulator. For example, the posteriors for $a_{\mathrm{sym}}^{\mathrm{NM}}$ $L_{\mathrm{sym}}^{\mathrm{NM}}$ match between the DKSVGP ($\mathrm{RMSE}=2.20~(\mathrm{MeV})$) and the DKSGP ($\mathrm{RMSE}=1.02~(\mathrm{MeV})$), but not the SVGP ($\mathrm{RMSE}=4.70~(\mathrm{MeV})$), while the posteriors for many other parameters are either unconstrained or loosely constrained and match between the DKSVGP and SVGP.

\FloatBarrier
\section{Conclusions}

In this work, we evaluated the performance of the sparse variational GP (SGP), the stochastic variational GP (SVGP), and their deep kernel counterparts the deep kernel sparse variational GP (DKSGP) and the deep kernal stochastic variational GP (DKSVGP). We created 64 versions of each of these 4 emulators by varying the number of inducing points and trained them uniformly within each emulator type. We evaluated performance on four metrics: prediction accuracy in the form of RMSE on a validation data set, prediction runtime, and inference results in the form of maximum likelihood parameter estimates and posterior distributions over the 12 UNEDF1 parameters under the emulator's predictive distribution and using the observed binding energies of 75 nuclides. We compared the predictive accuracy and runtime performance of these emulators to a multivariate GP+PCA emulator from prior work.\cite{schunck2020calibration} In addition, we compared maximum likelihood parameter estimates from the emulators with estimates that used the simulation directly during optimization.

We found that the approximate GPs frequently encountered instability during training--producing either invalid covariance matrix entries or non-symmetric or non-positive-definite covariance matrices for their predictive distributions--which made these models challenging to uniformly test as emulators. To make downstream comparisons of the emulator, we chose to use the emulator found during training that provided the best prediction accuracy on the validation set and additionally produced a valid, symmetric and positive definite covariance matrix for the predictive distribution at various points in the parameter space. Additionally, we found that the SGP emulators could get stuck in local optima instead of converging towards a global optima during training, revealing the gradient based optimization methods may not be sufficient for training SGP models.

None of these best-performing and stable approximate GP emulators surpassed the multivariate emulator in terms of prediction accuracy. We found that the best emulators in prediction accuracy and runtime were the SGP emulators with $m = 314$ inducing points. The DKSGP emulator with $m = 2$ inducing points provided prediction accuracy within a factor of $2$ of the best performing SGP with an order of magnitude improvement in runtime. In contrast, the worst predicting emulators were the SVGP and DKSVGP, achieving an order of magnitude worse performance in prediction RMSE, but with an up to one or two orders of magnitude improvement in prediction runtime, surpassing the multivariate emulator in runtime performance.

Only the SGP and SVGP emulators provided maximum likelihood parameter estimates that were broadly consistent with simulation-derived estimates from prior work.\cite{schunck2020calibration} Additionally, this work revealed that it can be challenging to find the global maximum likelihood estimate using gradient based optimizers, as all emulators provided inconsistent maximum-likelihood estimates for at least one parameter across repeated rounds of maximum likelihood estimation with the LBFGS optimizer. The DKSGP and DKSVGP emulators provided maximum likelihood parameter estimates that were inconsistent with prior work and broadly inconsistent with one another across the choice of number of inducing points. Additionally, the SGP and DKSGP emulators were the most useful for producing constrained posterior distributions over the UNEDF1 simulation parameters when comparing emulator predictions to observed binding energies.

The approximate GP emulators analyzed in this work show both utility and promise. While not providing prediction accuracy that is as good as prior work, these emulators can still be useful for inferring maximum likelihood parameter estimates and posterior distributions over the 12 UENDF1 parameters. The comparatively poor prediction accuracy was due to early stopping during training resulting from instabilities in the models. The fully trained approximate GP emulators, especially the DKSGP, were able to approach the multivariate emulator in terms of prediction accuracy. This indicates that there is additional potential not revealed by this work in the use of the DKSGP and SGP for emulating the UNEDF1 simulator if the issues of numerical stability can be resolved. Additionally, there is potential in the use of the SVGP and DKSVGP emulators in applications where emulator runtime is critical, since these emulators provided the fastest runtime over all and were almost an order of magnitude faster than emulators from prior work. Future work on this topic should consider how to modify approximate GP emulators or their training in order to avoid numerical instabilities that hinder their performance.

\acknowledgments % equivalent to \section*{ACKNOWLEDGMENTS}       
 
This material is based upon work supported by the U.S. Department of Energy, Office of Science, Office of Advanced Scientific Computing Research, and Department of Energy Computational Science Graduate Fellowship under Award Number DE-SC0019323. This work was supported by the U.S.\ Department of Energy, Office of Science, Offices of Advanced Scientific Computing Research and Nuclear Physics SciDAC programs under Contract number 89233218CNA000001 
and by the NUCLEI SciDAC project. This research used resources provided by the Darwin testbed at Los Alamos National Laboratory (LANL) which is funded by the Computational Systems and Software Environments subprogram of LANL's Advanced Simulation and Computing program (NNSA/DOE). This work was facilitated through the use of advanced computational, storage, and networking infrastructure provided by the Hyak supercomputer system at the University of Washington

This manuscript has been approved for release by the Los Alamos Classification Office and assigned number LA-UR-22-28858.

% References
\bibliography{report} % bibliography data in report.bib

\begin{thebibliography}{10}

\bibitem{kennedy2001bayesian}
Kennedy, M.~C. and O'Hagan, A., ``Bayesian calibration of computer models,''
  {\em Journal of the Royal Statistical Society: Series B (Statistical
  Methodology)}~{\bf 63}(3),  425--464 (2001).

\bibitem{higdon2004combining}
Higdon, D., Kennedy, M., Cavendish, J.~C., Cafeo, J.~A., and Ryne, R.~D.,
  ``Combining field data and computer simulations for calibration and
  prediction,'' {\em SIAM Journal on Scientific Computing}~{\bf 26}(2),
  448--466 (2004).

\bibitem{emulation_higdon_2008}
Higdon, D., Gattiker, J., Williams, B., and Rightley, M., ``Computer model
  calibration using high-dimensional output,'' {\em Journal of the American
  Statistical Association}~{\bf 103}(482),  570--583 (2008).

\bibitem{rasmussen2003gaussian}
Rasmussen, C.~E., ``Gaussian processes in machine learning,'' in [{\em Summer
  school on machine learning}{\nolinebreak\hspace{0.1em}]},   63--71, Springer
  (2003).

\bibitem{sparse_gp_titsias}
Titsias, M., ``Variational learning of inducing variables in sparse gaussian
  processes,'' in [{\em Artificial intelligence and
  statistics}{\nolinebreak\hspace{0.1em}]},   567--574, PMLR (2009).

\bibitem{hensman2013gaussian}
Hensman, J., Fusi, N., and Lawrence, N.~D., ``Gaussian processes for big
  data,'' {\em arXiv preprint arXiv:1309.6835}  (2013).

\bibitem{svgp_hensman}
Hensman, J., Matthews, A., and Ghahramani, Z., ``Scalable variational gaussian
  process classification,'' in [{\em Artificial Intelligence and
  Statistics}{\nolinebreak\hspace{0.1em}]},   351--360, PMLR (2015).

\bibitem{deep_kernel_wilson}
Wilson, A.~G., Hu, Z., Salakhutdinov, R., and Xing, E.~P., ``Deep kernel
  learning,'' in [{\em Artificial intelligence and
  statistics}{\nolinebreak\hspace{0.1em}]},   370--378, PMLR (2016).

\bibitem{mumpower2016impact}
Mumpower, M.~R., Surman, R., McLaughlin, G., and Aprahamian, A., ``The impact
  of individual nuclear properties on r-process nucleosynthesis,'' {\em
  Progress in Particle and Nuclear Physics}~{\bf 86},  86--126 (2016).

\bibitem{PhysRevC.85.024304}
Kortelainen, M., McDonnell, J., Nazarewicz, W., Reinhard, P.-G., Sarich, J.,
  Schunck, N., Stoitsov, M.~V., and Wild, S.~M., ``Nuclear energy density
  optimization: Large deformations,'' {\em Phys. Rev. C}~{\bf 85},  024304 (Feb
  2012).

\bibitem{schunck2020calibration}
Schunck, N., O’Neal, J., Grosskopf, M., Lawrence, E., and Wild, S.,
  ``Calibration of energy density functionals with deformed nuclei,'' {\em
  Journal of Physics G: Nuclear and Particle Physics}~{\bf 47}(7),  074001
  (2020).

\bibitem{mcdonnell2015uncertainty}
McDonnell, J., Schunck, N., Higdon, D., Sarich, J., Wild, S., and Nazarewicz,
  W., ``Uncertainty quantification for nuclear density functional theory and
  information content of new measurements,'' {\em Physical review letters}~{\bf
  114}(12),  122501 (2015).

\bibitem{higdon2015bayesian}
Higdon, D., McDonnell, J.~D., Schunck, N., Sarich, J., and Wild, S.~M., ``A
  bayesian approach for parameter estimation and prediction using a
  computationally intensive model,'' {\em Journal of Physics G: Nuclear and
  Particle Physics}~{\bf 42}(3),  034009 (2015).

\bibitem{gardner2018gpytorch}
Gardner, J., Pleiss, G., Weinberger, K.~Q., Bindel, D., and Wilson, A.~G.,
  ``Gpytorch: Blackbox matrix-matrix gaussian process inference with gpu
  acceleration,'' {\em Advances in neural information processing systems}~{\bf
  31} (2018).

\bibitem{paszke2017automatic}
Paszke, A., Gross, S., Chintala, S., Chanan, G., Yang, E., DeVito, Z., Lin, Z.,
  Desmaison, A., Antiga, L., and Lerer, A., ``Automatic differentiation in
  pytorch,'' (2017).

\bibitem{bingham2019pyro}
Bingham, E., Chen, J.~P., Jankowiak, M., Obermeyer, F., Pradhan, N.,
  Karaletsos, T., Singh, R., Szerlip, P.~A., Horsfall, P., and Goodman, N.~D.,
  ``Pyro: Deep universal probabilistic programming,'' {\em J. Mach. Learn.
  Res.}~{\bf 20},  28:1--28:6 (2019).

\bibitem{PhysRevC.82.024313}
Kortelainen, M., Lesinski, T., Mor\'e, J., Nazarewicz, W., Sarich, J., Schunck,
  N., Stoitsov, M.~V., and Wild, S., ``Nuclear energy density optimization,''
  {\em Phys. Rev. C}~{\bf 82},  024313 (Aug 2010).

\bibitem{PhysRevC.87.044320}
Erler, J., Horowitz, C.~J., Nazarewicz, W., Rafalski, M., and Reinhard, P.-G.,
  ``Energy density functional for nuclei and neutron stars,'' {\em Phys. Rev.
  C}~{\bf 87},  044320 (Apr 2013).

\bibitem{andrianakis2012effect}
Andrianakis, I. and Challenor, P.~G., ``The effect of the nugget on gaussian
  process emulators of computer models,'' {\em Computational Statistics \& Data
  Analysis}~{\bf 56}(12),  4215--4228 (2012).

\bibitem{tipping1999probabilistic}
Tipping, M.~E. and Bishop, C.~M., ``Probabilistic principal component
  analysis,'' {\em Journal of the Royal Statistical Society: Series B
  (Statistical Methodology)}~{\bf 61}(3),  611--622 (1999).

\bibitem{lbfgs}
Liu, D.~C. and Nocedal, J., ``On the limited memory bfgs method for large scale
  optimization,'' {\em Mathematical programming}~{\bf 45}(1),  503--528 (1989).

\bibitem{salimbeni2018natural}
Salimbeni, H., Eleftheriadis, S., and Hensman, J., ``Natural gradients in
  practice: Non-conjugate variational inference in gaussian process models,''
  in [{\em International Conference on Artificial Intelligence and
  Statistics}{\nolinebreak\hspace{0.1em}]},   689--697, PMLR (2018).

\bibitem{adam}
Kingma, D.~P. and Ba, J., ``Adam: A method for stochastic optimization,'' {\em
  arXiv preprint arXiv:1412.6980}  (2014).

\bibitem{nuts}
Hoffman, M.~D., Gelman, A., et~al., ``The no-u-turn sampler: adaptively setting
  path lengths in hamiltonian monte carlo.,'' {\em J. Mach. Learn. Res.}~{\bf
  15}(1),  1593--1623 (2014).

\bibitem{gelman1992inference}
Gelman, A. and Rubin, D.~B., ``Inference from iterative simulation using
  multiple sequences,'' {\em Statistical science} ,  457--472 (1992).

\end{thebibliography}
\bibliographystyle{spiebib} % makes bibtex use spiebib.bst

\end{document}